\documentclass[runningheads]{llncs}

\usepackage{graphicx}
\usepackage{algpseudocode}
\usepackage{algorithm}
\usepackage{mathtools}
\usepackage{cleveref}
\usepackage{enumitem}   
\usepackage{float}
\DeclareUnicodeCharacter{2212}{-}

\begin{document}

\title{XInsight: Revealing Model Insights for GNNs with Flow-based Explanations}

\titlerunning{XInsight}
%
\author{Anonymous for Double-Blind Review}
\author{Eli Laird\inst{1}\orcidID{0000-0002-0668-8745} \and
Ayesh Madushanka\inst{1}\orcidID{0000-0001-8151-2208} \and
Elfi Kraka\inst{1}\orcidID{0000-0002-9658-5626} \and
Corey Clark\inst{1}\orcidID{0000-0003-0129-9270}}
\authorrunning{Laird et al.}
%
\institute{Southern Methodist University, Dallas TX, USA
\email{\{ejlaird,amahamadakalapuwage,ekraka,coreyc\}@smu.edu}}

\maketitle             
%
\begin{abstract}

Progress in graph neural networks has grown rapidly in recent years, with many new developments in drug discovery, medical diagnosis, and recommender systems. While this progress is significant, many networks are `black boxes' with little understanding of the `what' exactly the network is learning. Many high-stakes applications, such as drug discovery, require human-intelligible explanations from the models so that users can recognize errors and discover new knowledge. Therefore, the development of explainable AI algorithms is essential for us to reap the benefits of AI.

We propose an explainability algorithm for GNNs called eXplainable Insight (XInsight) that generates a distribution of model explanations using GFlowNets. Since GFlowNets generate objects with probabilities proportional to a reward, XInsight can generate a diverse set of explanations, compared to previous methods that only learn the maximum reward sample. We demonstrate XInsight by generating explanations for GNNs trained on two graph classification tasks: classifying mutagenic compounds with the MUTAG dataset and classifying acyclic graphs with a synthetic dataset that we have open-sourced. We show the utility of XInsight's explanations by analyzing the generated compounds using QSAR modeling, and we find that XInsight generates compounds that cluster by lipophilicity, a known correlate of mutagenicity. Our results show that XInsight generates a distribution of explanations that uncovers the underlying relationships demonstrated by the model. They also highlight the importance of generating a diverse set of explanations, as it enables us to discover hidden relationships in the model and provides valuable guidance for further analysis.

\end{abstract}
\section{Introduction}

Graph neural networks (GNNs) have emerged as a popular and effective machine learning algorithm for modeling structured data, particularly graph data.
As GNNs continue to gain popularity, there is an increasing need for explainable GNN algorithms. Explainable AI refers to machine learning algorithms that can provide understandable and interpretable results. Explainable AI algorithms have the ability to uncover hidden relationships or patterns that deep learning models use in making their decisions. This means that researchers can use these methods to understand why a model arrived at a certain decision. In the case of GNNs, the need for explainability arises from the fact that they are often used in applications where the decision-making process needs to be transparent and easily understood by humans. For example, in the field of drug discovery, GNNs are used to predict the efficacy of a drug by analyzing its molecular structure \cite{zhang_graph_2022}. In this case, it is crucial to understand how the GNN arrived at its prediction, as it can have significant implications for patient health and safety. 

Explainable AI algorithms also uncover erroneous correlations in deep learning models. For instance, in a study by Narla et al. \cite{narla_automated_2018}, the researchers found that their model had incorrectly learned that images with rulers were more likely to be cancerous. The use of explainable AI methods helped them uncover this error and highlighted the need for methods that can explain the underlying relationships that deep learning models rely on to make predictions.

In response to this need, we propose a GNN explainability algorithm, \textit{eXplainable Insight (XInsight)}, that generates diverse model-level explanations using Generative Flow Networks (GFlowNets) \cite{bengio_flow_2021}. XInsight represents the first application of GFlowNets to explain graph neural networks. Unlike previous model-level algorithms, that only learn the maximum reward sample, XInsight generates objects with probabilities proportional to a reward. We demonstrate the effectiveness of XInsight by applying it to GNNs trained on two graph classification tasks: classifying mutagenic compounds with the MUTAG dataset and classifying acyclic graphs with a synthetic dataset. In our experiments, we demonstrate that XInsight's explanations for the MUTAG dataset \cite{morris_tudataset_2020} can be analyzed using data mining techniques and QSAR \cite{honma_improvement_2019} modeling to uncover hidden relationships in the model. For instance, when analyzing the compounds generated by XInsight, we found that they clustered by lipophilicity, which is a known correlate of mutagenicity. Our results demonstrate that XInsight generates a distribution of explanations that enables the discovery of hidden relationships in the model.

The key contributions of this paper are summarized below:
\vspace{-.66cm}

\begin{enumerate}[label=(\roman*)]
    \item We proposed eXplainable Insight (XInsight), an explainability algorithm for Graph Neural Networks (GNNs) that uses GFlowNets to generate a distribution of model explanations.
    \item We applied XInsight to explain two classification tasks, one of which was a newly open-sourced synthetic dataset, and the other was a real-world molecular compound dataset.
    \item We analyzed XInsight's generated explanations using a clustering method and chemical analysis tool, which helped us to discover important underlying patterns and relationships of the examined model. 
\end{enumerate}

\section{Related Work}

\subsection{Graph Neural Networks}

Graph neural networks (GNNs) have emerged as a popular deep learning technique to model structured data that can be represented as graphs. Unlike traditional neural networks that operate on structured data like images and sequences, GNNs operate on non-Euclidean data, such as social networks \cite{fan_graph_2019,tan_deep_2019}, chemical molecules \cite{bongini_molecular_2021,gilmer_neural_2017,gasteiger_directional_2022,zhang_graph_2022}, and 3D point clouds \cite{gao_vectornet_2020,sheng_graph-based_2022}. GNNs typically use a message-passing approach \cite{gilmer_neural_2017}, where the feature representations of nodes, edges, and the overall graph are iteratively updated by aggregating the features of their neighbors and combining them with the learned features from the previous step. This message-passing process is repeated for a fixed number of iterations or until convergence. Expanding upon traditional message-passing GNNs, many other GNN architectures have been proposed, such as Graph Convolutional Networks (GCNs) \cite{zhang_semi-supervised_2022} that use convolutional operations similar to Euclidean Convolutional Neural Networks, Graph Isomorphism Networks (GINs) \cite{xu_how_2019} that employ multilayer perceptrons to aggregate neighboring features, and Graph Attention Networks (GATs) \cite{velickovic_graph_2018} that apply an attention mechanism to weigh contributions of neighboring nodes/edges based on their importance. With the development of GNNs, we can now model and make predictions based on structured data in a way that was not possible before.

\subsection{Explaining Graph Neural Networks}

Graph neural networks (GNNs) are widely used in various domains such as drug discovery \cite{bongini_molecular_2021,gilmer_neural_2017,gasteiger_directional_2022,zhang_graph_2022}, recommendation systems \cite{ying_graph_2018,wu_self-supervised_2021,wang_neural_2019}, and medical diagnosis \cite{ahmedt-aristizabal_graph-based_2021,lu_weighted_2021,li_graph_2020}. However, as with other machine learning models, GNNs are often considered to be `black boxes', providing little insight into how they make predictions. Therefore, explainable AI algorithms for GNNs have gained increasing attention in recent years.

There are several approaches to developing explainable GNN algorithms that can conveniently be categorized as \textit{instance-level} and \textit{model-level} approaches.  Instance-level algorithms provide explanations for individual predictions of the GNN and include methods that utilize the gradients of the features to determine input importance, such as sensitivity analysis, Guided BP, and Grad-CAM \cite{baehrens_how_2009,springenberg_striving_2015,selvaraju_grad-cam_2020}, perturb inputs to observe changes in output as in GNNExplainer and PGExplainer \cite{gnnExplainer,pgExplainer}, and learn relationships between the input and its neighbors using surrogate models \cite{huang_graphlime_2020,vu_pgm-explainer_2020}. While there are several instance-level explainability methods for GNNs, there is still a lack of effective model-level explainability methods \cite{yuan_explainability_2022}. 

Model-level explanations help identify how the GNN approaches the task at hand, and can reveal patterns and structures that may not be immediately evident from the graph data alone. They also help identify when a model is not performing well on the given task, or when it is exhibiting unwanted behavior. Outside of the graph-learning world, input optimization is a popular model-level approach for image classification models, where the goal is to generate an image that maximizes the predicted class label \cite{nguyen_deep_2015,nguyen_plug_2017,mahendran_understanding_2015,dosovitskiy_inverting_2016,simonyan_deep_2014,olah_feature_2017}. In contrast, model-level explanations for GNNs have received relatively less attention. One of the most prominent model-level explainability methods for GNNs is XGNN (eXplainable Graph Neural Networks) \cite{yuan_xgnn_2020}. XGNN leverages reinforcement learning techniques to generate graphs that provide insights into how the GNN is making predictions. XGNN generates a graph explanation that maximizes a target prediction, thereby revealing the optimized class pattern learned by the model.

To gain more insight into the model, it is often necessary to analyze a diverse distribution of examples that cover different scenarios and edge cases. Furthermore, generating a distribution of explanations opens the door to applying statistical analysis and data mining techniques, such as dimensionality reduction or t-tests, to uncover hidden relationships in the data. For instance, in this paper we use dimensionality techniques to uncover clusters within XInsight explanations. When then used these clusters to verify that the model correctly learned a known correlation within the data. 

\vspace{-.1cm}
\subsection{XGNN}
\label{xgnn-section}
\vspace{-.1cm}

XGNN, which stands for eXplainable Graph Neural Networks, is a novel model-level explainability framework introduced by Yuan et al. in 2020 \cite{yuan_xgnn_2020}. The goal of XGNN is to generate a graph that maximizes a specific target class of a graph classification model. XGNN employs a reinforcement learning approach to iteratively build a graph using actions that add nodes or edges to the graph at each time step. During each time step, the model calculates the reward based on the probability of the target class, which encourages the algorithm to select actions that generate graphs of a particular class. This process is repeated until the model converges or until a maximum number of time steps is reached.

Like Graph Convolutional Policy Networks \cite{gcpn_leskovec}, XGNN learns a generator model using a policy gradient. The generator produces a graph that contains patterns that maximize the target class in question. In contrast to instance-level explainability methods that identify subgraphs that contribute to the model's output, XGNN focuses on the entire graph and the relationships between its nodes and edges. XGNN is currently the only model-level explanation method that has been proposed for GNNs, according to a recent survey \cite{yuan_explainability_2022}.

While XGNN is a powerful model-level explainability method for GNNs, it generates a single maximum reward explanation, which limits its ability to explain the full extent of the model's behavior. XGNN is also limited in its utility to discover hidden insights related to the classification task due to the inability to perform a more detailed analysis of the explanations, such as clustering. Due to these limitations, there is no way of directly comparing XGNN to techniques that generate a distribution of explanations, such as XInsight.

\subsection{GFlowNets}
\label{section:gflownet}

Generative Flow Networks (GFlowNets) are a type of generative model that generate a diverse set of objects by iteratively sampling actions proportional to a reward function \cite{bengio_flow_2021} \cite{bengio_gflownet_2022}. The objective of GFlowNets is to learn to sample from a distribution of diverse and high-reward samples instead of generating a single sample to maximize a reward function. GFlowNets can be viewed as Markov Decision Processes (MDP) represented by a directed acyclic graph (DAG), where the edges represent the actions that can be taken in the states. The flows coming into a state represent the actions that can be taken to reach that state, while the flows leaving a state represent the actions that can be taken in that state to reach the next state. The DAG is traversed iteratively by sampling flows, which generates a flow trajectory that ends when a terminal state is reached. The flow entering a terminal state is the total flow of the trajectory and is equal to the reward function assigned to that state.

\vspace{-.1cm}
\subsubsection{Trajectory Balance Objective}
\label{subsection:trajectory-balance-objective}

In \cite{malkin_trajectory_2022}, Malkin et al. introduced the \textit{Trajectory Balance Constraint}, shown in Equation~\ref{eq:trajectory-balance-constraint}, which ensures that the flow of the trajectory leading to a state is equal to the flow of the trajectory leaving that state and terminating at a terminal state. Satisfying this constraint allows the GFlowNet to sample objects with probability proportional to its reward.

\begin{equation}
    \label{eq:trajectory-balance-constraint}
    Z \prod_{t} P_F(s_{t+1}|s_t) = R(\tau)\prod_t P_B(s_t|s_{t+1})
\end{equation}

where $Z$ is the `total flow'. The right side of Equation~\ref{eq:trajectory-balance-constraint} represents the fraction of the total reward going through the trajectory, while the left represents the fraction of the total flow going through the trajectory. This constraint can be turned into the \textit{Trajectory Balance Objective} \cite{malkin_trajectory_2022} for training a GFlowNet, shown below:

\begin{equation}
    \label{eq:trajectory-balance-objective}
    L_{TB}(\tau) = \left(\log \frac{Z \prod_t P_F(s_{t+1}|s_t)}{R(\tau)\prod_t P_B(s_t|s_{t+1})} \right)^2
\end{equation}

\subsubsection{Applications of GFlowNets}
\label{subsection:gflownet-applications}

GFlowNets have been applied to many generative applications, including molecular sequence generation \cite{bengio_flow_2021,jain_biological_2022,malkin_trajectory_2022} and MNIST image generation \cite{zhang_generative_2022}. And due to their ability to generate diverse samples, GFlowNets trained to generate model explanations, as in XInsight, provide the machine learning user a greater breadth of human-readable explanations of what their models are learning from the data.

\section{eXplainable Insight (XInsight)}

\subsection{Explaining Graph Neural Networks}

Graph classification networks can be difficult for humans to interpret since graph structures can be less intuitive to humans compared to visual features which humans are naturally equipped to interpret. Therefore when seeking to understand a graph classification model, a quality explainability algorithm should take advantage of the natural pattern matching capabilities of its human users by producing concise explanations that highlight patterns that are easily interpreted by humans. To make it even easier for its users, a quality explainability algorithm should produce a distribution of explanations in order to provide the user with multiple perspectives into the model; however, most algorithms to date lack one or both of these qualities.

XInsight not only produces concise explanations that highlight important patterns but also produces a distribution of explanations that allows the user to develop a more robust understanding of what the examined model is learning from the data. XInsight trains a GFlowNet to generate a diverse set of model-explanations for a graph classification model. The explanations that XInsight generates highlight general patterns that the classification model attributes to specified target class in question.

In the context of model explanations as a whole, the explanations that XInsight 
produces are particularly useful for discovering relationships within the trained model. For example, they can help determine if a model incorrectly associates an artifact in the data with the target class, like rulers with skin cancer as discussed in \cite{narla_automated_2018}. XInsight empowers users to do this by generating a distribution of explanations, which can then be passed through traditional data mining techniques, such as clustering, to uncover what the model is learning from the data. 

\subsection{Generating Graphs with XInsight}

XInsight employs a GFlowNet that it is trained to generate graphs with probabilities proportional to their likelihood of belonging to a target class. Specifically, the GFlowNet generates a graph by iteratively sampling actions that determine whether to add a new node or edge to the existing structure. It is important to note that the likelihood of a sample belonging to a particular class is defined by the trained model that is being explained. Therefore, the distribution of generated samples is dependent upon the trained model and not the true class distribution, which in the context of explaining a trained model is desirable since the goal is to understand the model itself.

\subsubsection{Action Space} The action space, $\mathcal{A}$, is split into two flows: the first selecting a starting node and the second selecting the ending node. The starting node is selected from the set of nodes $N$ in the current incomplete graph $G_t$. The ending node is selected from the union of the same $\mathcal{N}$, excluding the starting node and a set of building blocks $\mathcal{B}$. Together, the starting and ending nodes form the combined action $\mathcal{A}(n_s, n_e)$ sampled from the forward flow $P_F$. Taking this action generates a new graph $G_{t+1}$ as shown below:

\begin{equation}
    \label{eq:forward-flow}
    G_{t+1} \sim P_F(\mathcal{A}(n_s, n_e) | G_t)
\end{equation}

\begin{equation}
    \label{eq:starting-policy}
    p_{start}(n_s \in N | G_t)
\end{equation}

\begin{equation}
    \label{eq:ending-policy}
    p_{end}(n_e \in N\cup\mathcal{B}; n_e \neq n_s | G_t)
\end{equation}

\subsection{Proxy}

The proxy $f$ in classical GFlowNets is used to generate the reward for a generated object. For example, in \cite{bengio_flow_2021} Bengio et al. used a pretrained model as their proxy to predict the binding energy of a generated molecule to a protein target. In XInsight, we use the model to be explained as the proxy since the generated objects are treated as explanations of the model.

\subsubsection{Reward} \label{subsubsection:reward} The reward in XInsight guides the underlying GFlowNet to generate graphs that explain the proxy. In XInsight, we define the reward as the proxy's predicted probability that the generated graph belongs to the target class $c$, as shown in Equation~\ref{eq:reward}. To encourage the generation of objects explaining the target class, we define the reward to be zero if the generated object is classified as the opposite class. In addition, we add a scalar multiplier $\alpha$ to magnify the reward for the target class, where $\alpha > 0$.

\begin{equation}
\label{eq:reward}
    R(G_t) = 
    \begin{cases}
        \alpha * softmax(f(G_t)) & \text{if}\; argmax\{f(G_t)\} = Target\; Class\\
        0 & \text{if} \; argmax\{f(G_t)\} \neq Target\; Class
    \end{cases}
\end{equation}

\subsection{Training XInsight}

We train XInsight using the trajectory balance objective, following \cite{malkin_trajectory_2022}. We define the trajectory balance objective for a complete graph $G$ generated over a trajectory $\tau$ actions in Equation~\ref{eq:trajectory-balance-objective-xi}. 

\begin{equation}
    \label{eq:trajectory-balance-objective-xi}
    L_{TB}(G) = \left(\log \frac{Z \prod_t^{\tau} P_F(G_{t+1}|G_t)}{R(G_t)\prod_t^{\tau} P_B(G_t|G_{t+1})} \right)^2
\end{equation}

The training loop consists of sampling trajectories (i.e. generating graphs), calculating forward and backward flows and the reward, and updating the underlying GFlowNet parameters until convergence. We highlight the in-depth steps of XInsight's training loop in Algorithm~\ref{alg:xi-training}. 

For every epoch in the training loop, we start by initializing the GFlowNet and creating an initial graph $G_0$. Then we generate a graph by iteratively sampling actions from the forward policy $P_F$ that add nodes or edges to the graph at each step $G_t$. Once a trajectory is complete, either by the reaching the $MAX\_ACTIONS$ limit or by sampling a $stop$ action, the reward is computed for $G_t$ using the $Proxy$. Finally, we calculate the trajectory balance loss, update the GFlowNet's parameters and repeat.

\begin{algorithm}
\caption{XInsight Training Loop}\label{alg:xi-training}
    \textbf{Input}: $EPOCHS$, $Proxy(\cdot)$, $TARGET\_CLASS$, $MAX\_ACTIONS$
    \begin{algorithmic}
        \State $XI(\cdot; \theta) \gets$ GFlowNet
        \For{epoch in $EPOCHS$}
            \State $actions \gets 0$
            \State $G_t \gets G_0$ \Comment{Initialize new graph}
            \State $\tau \gets \emptyset + \{G_t\}$
            \Repeat
                \State $P_F, P_B \gets XI(G_t; \theta)$ \Comment{Generate flows}
                \State $n_s, n_e \sim P_F$ \Comment{Sample start \& end node}
                \If {$n_s = stop$}
                    \State $STOP \gets True$ \Comment{Stop if stop action sampled}
                \EndIf
                \State $G_{new} \gets$ $\mathcal{T}(n_s, n_e)$ \Comment{Add node/edge}
                \State $P_F, P_B \gets XI(G_{new}; \theta)$ \Comment{Recompute flows}
                \State $G_t = G_{new}$
                \State $\tau \gets \tau + \{G_t\}$ \Comment{Append $G_t$ to trajectory}
                \State $actions ++$
            \Until $actions > MAX\_ACTIONS$ or $STOP$
            \State $Reward = softmax(Proxy(G_t))_{TARGET\_CLASS}$ \Comment{Calculate reward}
            \State $\theta \gets \theta - \eta \nabla Loss_{TB}(Reward, log_{Z}, \tau)$ \Comment{Update parameters}
        \EndFor
    \end{algorithmic}
\end{algorithm}

\section{Experiment Design}
\label{section:experiment-design}
\vspace{-.2cm}

\subsection{Datasets}
\label{subsection:datasets}

The Acyclic Graph dataset includes 2405 synthetically generated graphs labeled as either acyclic or cyclic. We generated graphs using graph generation functions from the NetworkX software package \cite{networkx}. To improve the diversity of the dataset, we trained a GFlowNet with a brute-force cycle checker as a reward function to generate acyclic and cyclic graphs to add to the dataset. The code used to generate this dataset can be found in \cite{acyclic-graph}.

The MUTAG dataset \cite{morris_tudataset_2020}, included in Pytorch Geometric, contains 188 graphs representing chemical compounds used in an Ames test on the S. Typhimurium TA98 bacteria with the goal of measuring the mutagenic effects of the compound. This dataset was used in a study to measure the correlation between the chemical structure of the compounds and their mutagenic activity \cite{debnath_structure-activity_1991}. The nodes and edges in the graphs in MUTAG represent 7 different atoms (Carbon, Nitrogen, Oxygen, Fluorine, Iodine, Chlorine, and Bromine) and their chemical bonds. In the graph learning community, the dataset is used as a benchmark dataset for graph classification models labeling each graph as `Mutagenic' or `Non-Mutagenic'.

\subsection{Verifying XInsight's Generative Abilities Setup}
\label{subsection:acyclic-experiment}

To validate that XInsight can generate graphs belonging to a target class, we trained XInsight to generate acyclic graphs because of their simple and human-interpretable form. For the proxy, we trained a graph convolutional neural network (GCN) to classify acyclic graphs using the Acyclic Graph dataset and node degree as the node features, achieving $99.58\%$ accuracy. This GCN is composed of three graph convolutional layers (GCNConv) with 32, 48, 64 filters, respectively, a global mean pooling layer, and two fully connected layers with 64 and 32 hidden units. We also used dropout and the ReLU activation function. The GFlowNet was also a GCN made up of three GCNConv layers with 32, 64, and 128 filters, two fully connected layers with 128 and 512 hidden units, and a scalar parameter representing $log(Z)$ from the reward function, see Section~\ref{subsubsection:reward}. The building blocks used for action selection consisted of a single node of degree 1.

\subsection{Revealing MUTAG Relationships Setup}
\label{subsection:mutag-experiment}

Due to their highly qualitative nature, there is no established method for evaluating model-level explanation methods for graphs, particularly for methods that generate a distribution of explanations. Despite this barrier, we demonstrate XInsight's explanatory abilities by applying it to the task of knowledge discovery within the mutagenic compound domain. Particularly, we evaluated XInsight for its ability to uncover meaningful relationships learned by a graph neural network trained to classify mutagenic compounds and verify that these relationships exist in the ground truth data.

\begin{figure}[!ht]
    \centering
    \includegraphics[scale=0.25]{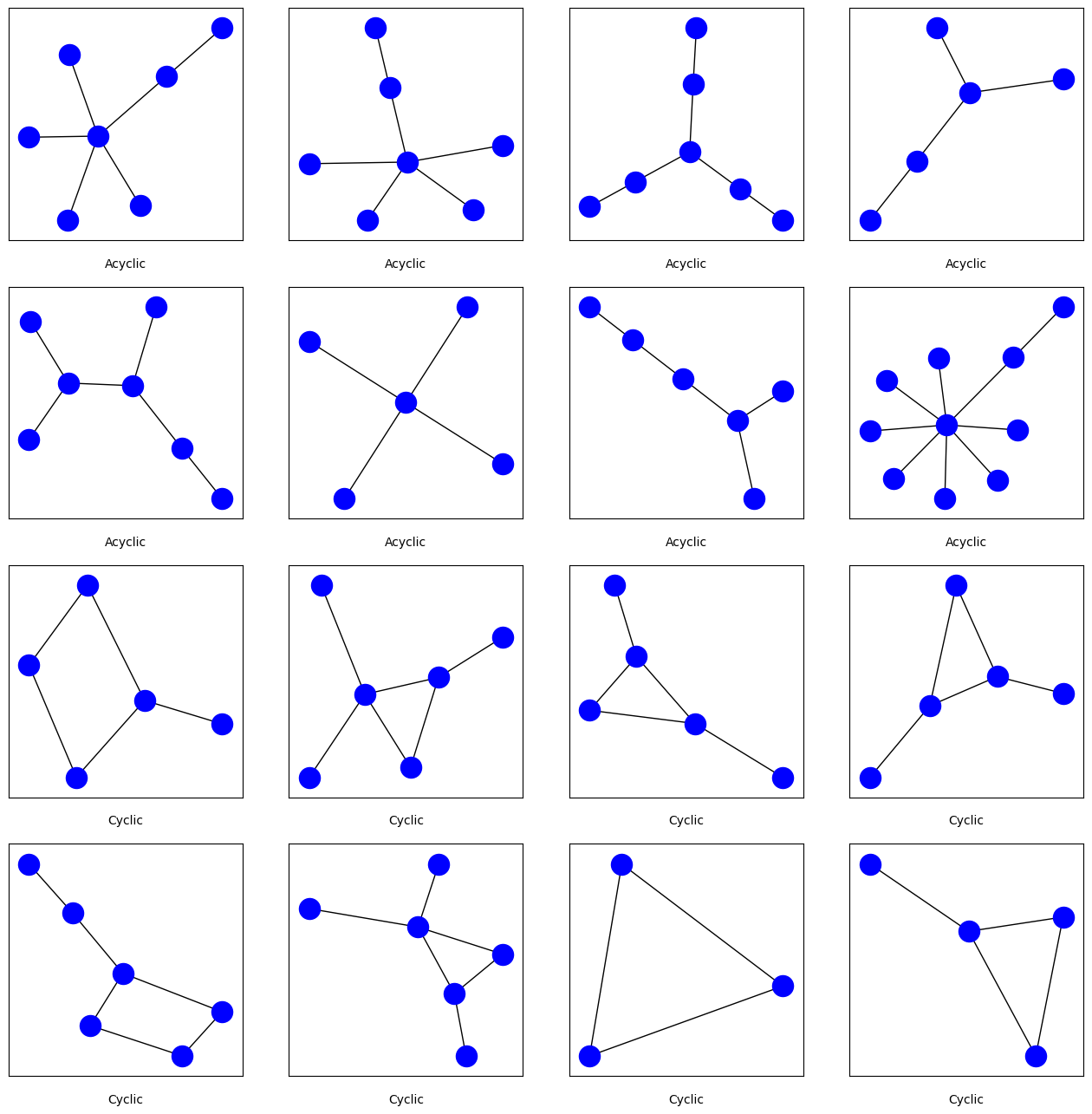}
    \caption{Generated graphs (8 with cycles and 8 without cycles) to verify XInsight's ability to generate graphs of a specified target class.}
    \label{fig:acyclic-explanations}
\end{figure}

For the proxy, we trained a graph convolutional neural network (GCN) to classify mutagenic compounds using the MUTAG dataset. Following \cite{yuan_xgnn_2020}, we used node features which were seven-dimensional one-hot encoded vectors encoding the seven different atoms in the dataset. The architecture of this GCN mirrored that used for the Acyclic classification task, with the addition of another GCNConv layer with 64 filters and LeakyReLU as the activation function. With this architecture, we achieved $89\%$ accuracy on the MUTAG classification task. The GFlowNet architecture was also the same as the one used for the Acyclic classification task, except we used seven nodes representing the different atoms as the building blocks. The initial graph $G_0$, used in training the GFlowNet, was set to a single node graph with the feature value set to carbon, as in \cite{yuan_xgnn_2020}. 

\section{Results}
\label{section:results}

\subsection{Verifying XInsight's Generative Abilities Using The Acyclic Dataset}
\label{subsection:acyclic-results}

To verify that XInsight is capable of generating graphs of a particular class defined by a classification model, we conducted an experiment in which we trained XInsight to generate graphs from a graph convolutional network, previously trained on the Acyclic Graphs dataset. This synthetic dataset contains two classes, acyclic and cyclic, and is described in detail in Section~\ref{subsection:datasets}.

Following XInsight training, we generated a distribution of 16 graphs (8 acyclic and 8 cyclic), shown in Figure~\ref{fig:acyclic-explanations}. The results of the experiment indicate that XInsight is indeed capable of generating acyclic graphs, which is consistent with the nature of the dataset. This provides evidence that XInsight is capable of generating graphs guided by the predictions of a simple classification model.

\begin{figure}[!t]
    \centering
    \includegraphics[scale=0.25]{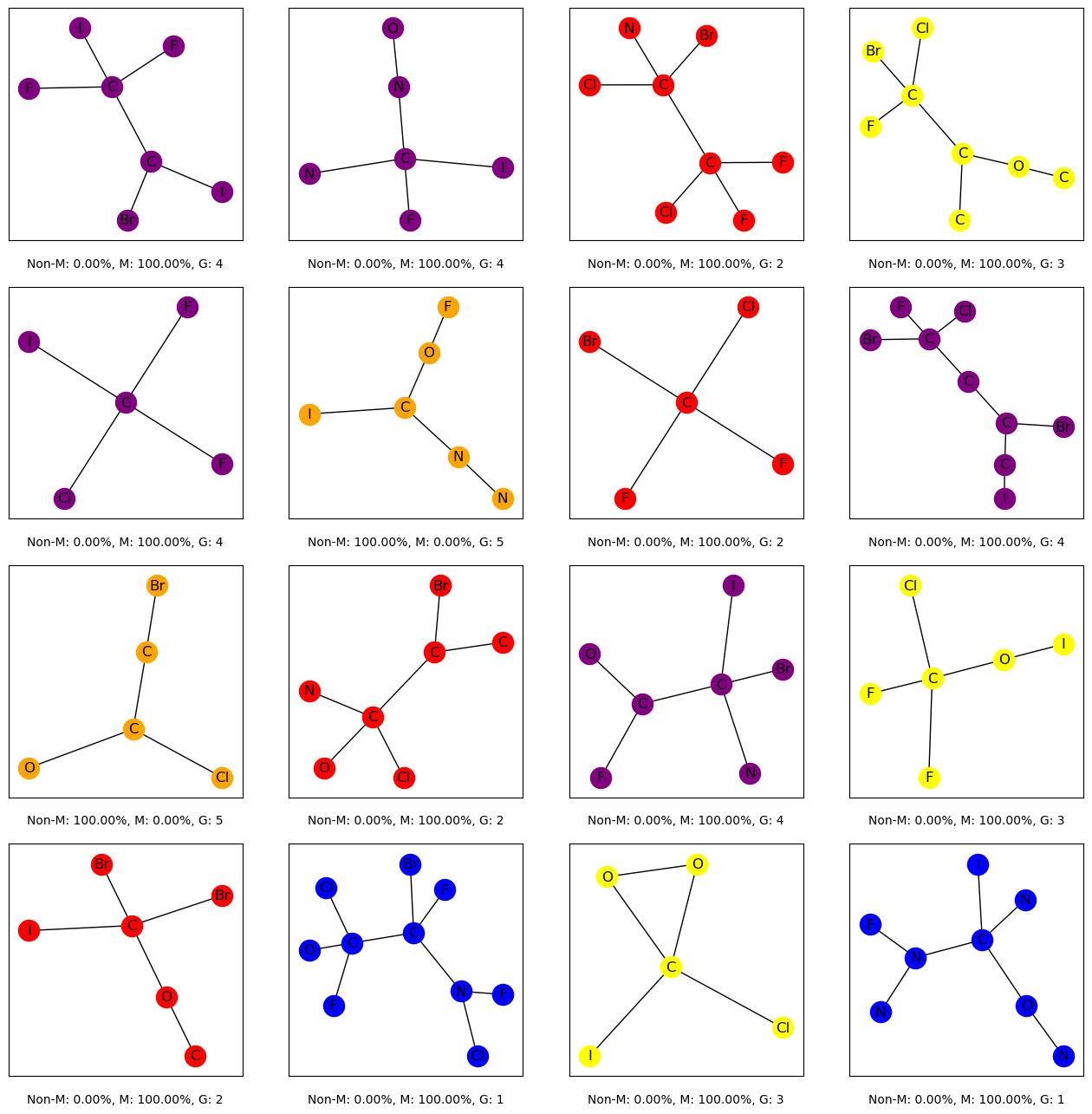}
    \caption{Distribution of explanations for the \textit{Mutagenic} classifier generated by the trained XInsight model, with MUTAG class probabilities according to the trained proxy. Colors represent UMAP clusters of graph embeddings for the generated compounds. Blue: Group 1, Red: Group 2, Yellow: Group 3, Purple: Group 4, Orange: Group 5.}
    \label{fig:mutagenic-explanations}
\end{figure}

\subsection{Revealing Distinct Relationships Learned By The MUTAG Classifier}

\subsubsection{Generating Explanations} In our second experiment, we trained XInsight to explain a GCN trained on the MUTAG dataset. Our objective was to uncover hidden patterns and relationships that the trained GCN classifier associates with the mutagenic class. To achieve this, we used XInsight to generate a distribution of 16 compounds, illustrated in \textit{Figure}~\ref{fig:mutagenic-explanations}, and then fed the generated graphs through the trained GCN to produce graph embeddings. In order to visualize the 32-dimensional graph embeddings we used the UMAP dimensionality reduction algorithm, which preserves global and local structure of the data \cite{mcinnes_umap_2020}, to project the embeddings onto a 2-dimensional plane. From this visualization, we identified five distinct groupings of compounds that we hypothesize group by an unknown factor related to mutagenicity. To uncover the factor behind these groupings, we continued our analysis by analyzing the chemical properties of the generated compounds using QSAR modeling \cite{honma_improvement_2019}.

\label{subsection:mutag-results}
\begin{figure}[!ht]
    \centering
    \includegraphics[scale=0.55]{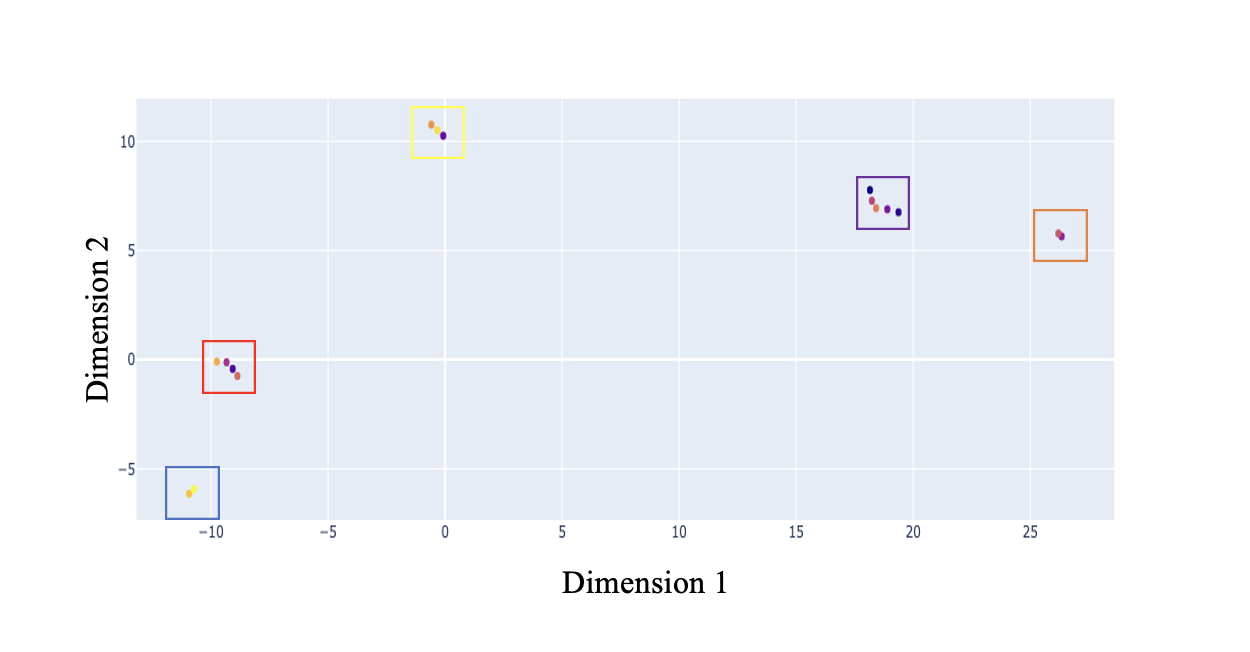}
    \caption{Generated graph embeddings projected onto 2-dimensional plane using UMAP. UMAP was fit using the cosine similarity metric, 2 neighbors, and a minimum distance of 0.1.}
    \label{fig:umap-clusters}
\end{figure}

\subsubsection{Knowledge Discovery} Quantitative Structure-Activity Relationship (QSAR) modeling is a well-established methodology that is used to differentiate between mutagenic and non-mutagenic compounds, which have been identified by the Ames test \cite{honma_improvement_2019}. Various features of typical drugs, such as lipophilicity, polarizability, hydrophilicity, electron density, and topological analysis, have been utilized in the literature to establish QSAR models for mutagenicity \cite{tuppurainen_frontier_1999}. Among these features, lipophilicity has been identified as a major contributing factor for mutagenicity, as it facilitates the penetration of lipophilic compounds through cellular membranes. 

\begin{figure}[!ht]
    \centering
    \includegraphics[scale=0.25]{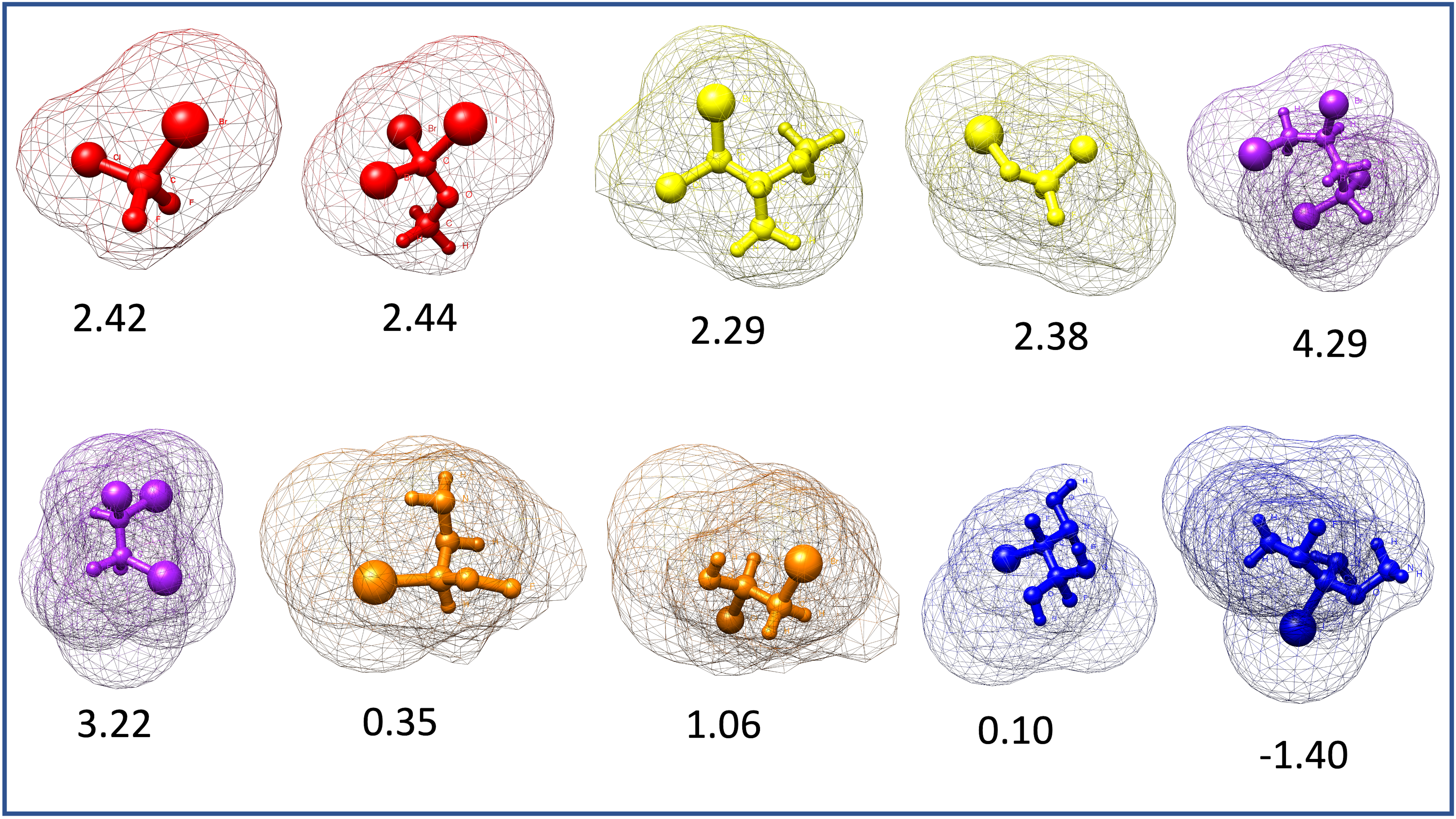}
    \caption{Lipophilicity calculations for 10 of the clustered compounds generated by XInsight using the XLOGP3 method. Surface mesh and 3 dimensional structures were generated by Chimera visualization software \cite{pettersen_ucsf_2004}.}
    \label{fig:lipophilicity}
\end{figure}

To establish a relationship between the clusters of compounds generated by XInsight and their mutagenicity, we calculated the lipophilicity of all the generated structures using the XLOGP3 method \cite{cheng_computation_2007}, samples shown in \ref{fig:lipophilicity}. This method has been shown to provide reliable results that are comparable to those obtained using the calculation of the octanol water partition coefficient for $log P$ \cite{viana_effect_2021}. It is essential to note that we added hydrogens to O (-OH) and N (-NH2) groups to represent the aqueous environment within the human body, since hydrogen atoms were not included in the building blocks for the MUTAG dataset. 

\begin{figure}
    \centering
    \includegraphics[scale=0.49]{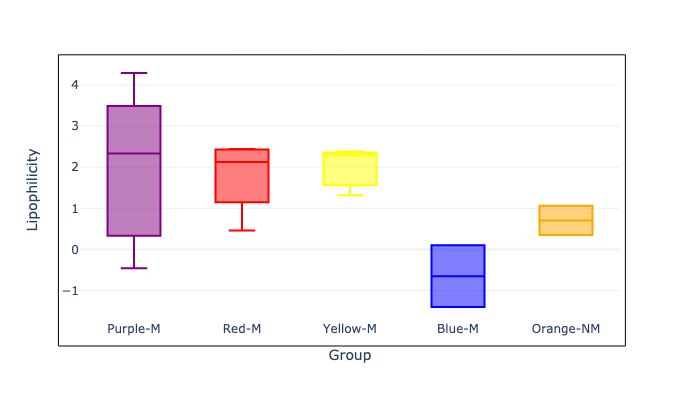}
    \caption{XLOGP3 Lipophilicity values for UMAP clustered compounds colored by group with classified labels, Mutagenic: \textit{M}, Non-Mutagenic: \textit{NM}.}
    \label{fig:lipophilicity-grouped}
\end{figure}

In \textit{Figure}~\ref{fig:lipophilicity-grouped} we see that in general the lipophilicity value is higher for the generated mutagenic compounds compared to the non-mutagenic compounds. We observed that the highest lipophilicity was associated with compounds of Group 4 (purple), followed by those of Group 2 (red) and Group 3 (yellow). The purple cluster exhibited significant differences in lipophilicity when compared to the red and yellow clusters, which explains why purple is a distinct cluster. However, groups 1 (blue) and 5 (orange) showed lower levels of lipophilicity values but still exhibited significant differences. Thus, lipophilicity appears to be a factor related to the mutagenicity of these compounds. 

\subsubsection{Knowledge Verification}
To verify that the discovered relationship between lipophilicity and mutagenicity is valid, we randomly sampled 32 compounds from the MUTAG dataset, with 16 compounds for each class, and calculated lipophilicity for each. We then performed a t-test to determine whether there is a statistically significant difference in lipophilicity for mutagenic and non-mutagenic compounds. In Table \ref{tab:ground-truth-tests} we show a statistically significant difference between the mean lipophilicity values for the mutagenic and non-mutagenic classes, thus verifying that the relationship uncovered using XInsight's generated distribution is a true relationship exhibited in the training data. Additionally, this shows how XInsight can be used to discover knowledge about the model.

\subsubsection{Further Insights} The distribution of explanations provided us with another significant insight, which is that the compounds in Group 1 (blue) and Group 5 (orange) have low lipophilicity, even though Group 1 is classified as mutagenic and Group 5 as non-mutagenic. This raises two possible assumptions: first, the classifier might be incorrectly classifying compounds that are similar to those in Groups 1 as mutagenic, or second, there might be another underlying factor that is responsible for the hydrophilic nature of these compounds. Furthermore, as mentioned earlier, lipophilicity is not the only factor determining the mutagenicity of the compounds. To explain the clustering of the Group 1, additional quantum-mechanical calculations are necessary.

This analysis underscores the considerable advantages of generating a distribution of explanations, as opposed to a single explanation that maximizes the reward. By having a distribution of explanations, we can uncover hidden insights into what the classification model associates with the target class. Without a distribution of explanations, we are restricted in the types of analysis we can perform to more effectively explain the model being examined. 

\begin{table}[hbp]
    \centering
    \caption{\textit{t-test} results showing a statistically significant difference between Mutagenic and Non-Mutagenic lipophilicity values for 32 randomly sampled compounds from the MUTAG dataset, $\alpha = 0.05$.}
    \label{tab:ground-truth-tests}
    \begin{tabular}{|c|ccc|}
        \hline
         & Mutagenic & Non-Mutagenic & \\
        \hline
        Mean & 4.1444 & 2.1750 & \\
        \hline
        Variance & 0.6812 & 1.3963 & \\
        \hline
        \multicolumn{1}{|c|}{t-statistic} & \multicolumn{2}{c}{-5.2917} & \\
        \hline
        \multicolumn{1}{|c|}{p-value} & \multicolumn{2}{c}{0.00001022} & \\
        \hline
    \end{tabular}
\end{table}

\section{Conclusion}

In this paper, we proposed XInsight, an explainability algorithm for graph neural networks, that generates a diverse set of model explanations using Generative Flow Networks. Our approach is designed to provide human-understandable explanations for GNNs that uncover the hidden relationships of the model. We demonstrated the effectiveness of XInsight by generating explanations for GNNs trained for two graph classification tasks, including the classification of acyclic graphs and the classification of mutagenic compounds. Our results indicate that XInsight uncovers underlying relationships and patterns demonstrated by the model, and provides valuable guidance for further analysis.

Our findings emphasize the importance of generating a diverse set of explanations, as it enables us to discover hidden relationships in the model and identify important features in the data. Furthermore, we show that the generated explanations from XInsight can be used in combination with data mining and chemical analysis methods to uncover relationships within the model. For instance, we analyzed the generated compounds from XInsight using QSAR modeling, and we observe that XInsight generates compounds that cluster by Lipophilicity, a known correlate of mutagenicity.

Overall, XInsight provides a promising direction for developing explainable AI algorithms for graph-based applications, with implications for many real-world domains. We believe that XInsight has the potential to make a significant impact in various real-world domains, particularly in high-stakes applications, such as drug discovery, where interpretability and transparency are essential.

\vspace{1cm}

%
%
%
\bibliographystyle{splncs04}
\bibliography{paper}

\end{document}